\documentclass[10pt,letterpaper,twocolumn]{article}

\usepackage[pagenumbers]{cvpr} 

\definecolor{cvprblue}{rgb}{0.21,0.49,0.74}
\usepackage[pagebackref,breaklinks,colorlinks,allcolors=cvprblue]{hyperref}
\usepackage{bm}
\usepackage{amsmath}
\usepackage{graphicx}
\usepackage[linesnumbered, ruled, lined]{algorithm2e}
\usepackage{multirow}
\usepackage{placeins}
\usepackage{utfsym}
\usepackage{array}
\usepackage{cuted}      
\usepackage{caption} 
\usepackage{float}
\usepackage{xcolor}
\usepackage[most]{tcolorbox}
\usepackage{booktabs}
\usepackage{multirow}
\usepackage[table]{xcolor}

\usepackage{algorithmic}
\usepackage{amssymb}

\DeclareFixedFont{\ttb}{T1}{txtt}{bx}{n}{8} 
\DeclareFixedFont{\ttm}{T1}{txtt}{m}{n}{8}  

\usepackage{color}
\definecolor{deepblue}{rgb}{0,0,0.5}
\definecolor{deepred}{rgb}{0.6,0,0}
\definecolor{deepgreen}{rgb}{0,0.5,0}

\def\paperID{6148} 
\def\confName{CVPR}
\def\confYear{2025}

\title{ProxyBuild: Text-Guided Structured 3D Building Generation with Mesh-Anchored Procedural Proxies}

\author{
Xiang Tang$^{1,2}$ \quad
Ruotong Li$^{2}$ \quad
Xiaopeng Fan$^{3,2,4}$ \\
$^1$Harbin Institute of Technology, Shenzhen \quad $^2$Pengcheng Laboratory \\ 
$^3$Harbin Institute of Technology \quad $^4$Harbin Institute of Technology, Suzhou Research Institute\\
}

\begin{document}

\twocolumn[{
\renewcommand\twocolumn[1][]{#1}
\maketitle

\begin{center}
    \includegraphics[width=\textwidth]{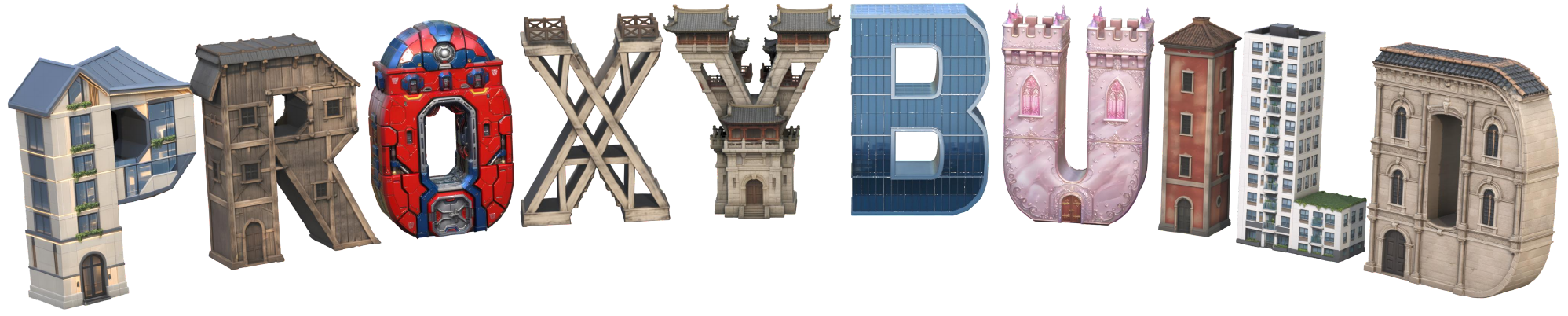}
    \captionof{figure}{ProxyBuild can process geometric shells of arbitrary shapes that satisfy pre-partitioning requirements, thereby generating structurally clear and stylistically diverse 3D buildings. This generative paradigm, tightly coupling macroscopic topological shells with parametric components, provides high-quality and editable 3D assets for downstream applications such as game development and digital twins.} 
    \label{teaser}
\end{center}
}]

\begin{abstract}
Text-guided 3D building generation holds tremendous application potential, yet existing generative models typically output inseparable single meshes or non-interactive rendered representations. While procedural modeling can generate editable buildings with hierarchical structures, rule authoring is laborious, and even with the aid of large language models (LLMs), it remains challenging to effectively solve procedural rules under geometric constraints. In this paper, we propose ProxyBuild, a novel hybrid framework for structured building generation. We introduce the Mesh-Anchored Procedural Proxy (MAPP) as a novel intermediate representation, which tightly anchors building components onto geometric shells, thereby decoupling the generation task into two phases: proxy prediction and proxy-to-asset instantiation. First, we construct a building dataset with MAPP annotations to train our designed face-edge bigraph encoder. By explicitly modeling the feature interactions of topological elements on heterogeneous mesh graphs, this encoder accurately infers the semantic roles of faces and edges. Subsequently, conditioned on textual styles and attribute parameters parsed by LLMs, we accomplish high-precision asset retrieval and assembly by integrating a spatial placement logic with hard constraints. Extensive experiments show that ProxyBuild not only significantly mitigates common issues in building generation such as over-smoothing, component collisions, and structural corruptions, but also accurately parses semantic-free shells from diverse sources. Outperforming prior baselines across various metrics, our method can robustly generate structurally clear, detail-rich, and post-editable 3D buildings from text, thereby providing a reliable and interactive content foundation for downstream applications such as virtual reality and digital twins.
\end{abstract}
\section{Introduction}

Automated 3D building generation plays a pivotal role in fields such as computer graphics, game development, and digital twins. Early research primarily relied on procedural content generation (PCG) techniques \cite{christiansen2012generic, hu2021extended}, formulating shape grammars to construct buildings with high-quality geometry and hierarchical structures. However, PCG heavily depends on expert prior knowledge, and the hard-coded nature of rules makes it difficult to flexibly accommodate diverse, high-level semantic natural language instructions from users. An intuitive approach is to introduce LLMs to understand text and automatically generate or drive PCG rules. Nevertheless, LLMs inherently lack precise 3D spatial perception and continuous numerical reasoning capabilities, making it difficult to solve PCG under complex geometric constraints.

On the other hand, thanks to recent breakthroughs in diffusion models and 3D representations, text-guided 3D building generation \cite{xiang2025structured, huang2026majutsucity} has witnessed remarkable progress. Although these generative paradigms substantially broaden the diversity of architectural forms, the generated objects are typically implicit neural fields (e.g., NeRF, 3DGS) or monolithic triangle meshes. These representations lack structured hierarchical information, failing to satisfy the demands for interactive editing and asset reuse in downstream pipelines. To address this issue, recent hybrid generative frameworks, exemplified by BuildingBlock \cite{huang2025buildingblock}, attempt to introduce structural priors by combining spatial bounding boxes with PCG. Nevertheless, mere stacking and loose arrangement based on bounding boxes often overlooks fine-grained architectural boundary constraints, which is prone to causing geometric artifacts such as floating components and mesh interpenetration during the subsequent component instantiation stage.

How can we break the barrier between unstructured geometric representations and parametric component entities to achieve physically plausible yet highly flexible building generation? Our core insight is that the generated structure of a building should be directly and tightly anchored to the topology (i.e., faces and edges) of its underlying base geometry. Driven by this philosophy, we propose ProxyBuild, a novel structured hybrid generative method for 3D buildings. This framework introduces the MAPP as an intermediate representation, and decouples the entire building generation task into two distinct phases.

\begin{figure*}
    \centering
    \includegraphics[width=\textwidth]{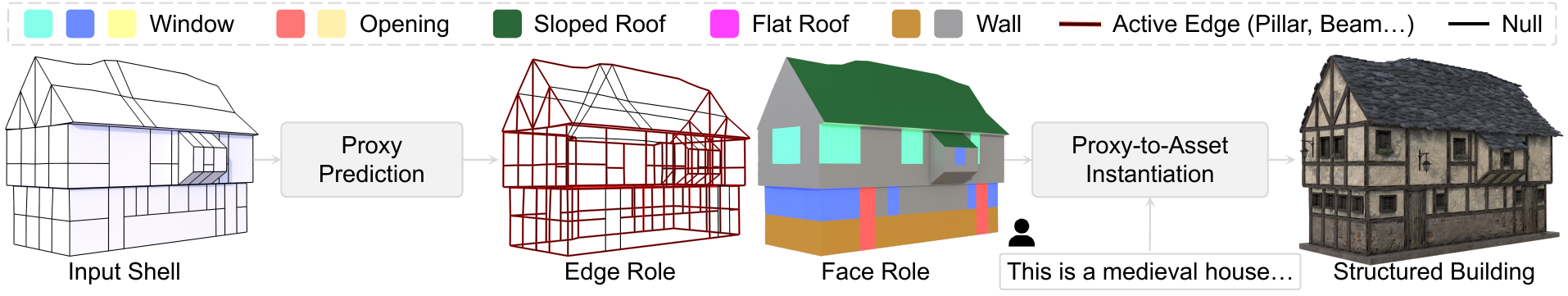}
    \caption{The generative pipeline of ProxyBuild. Given a text prompt and an input shell, the framework proceeds in two stages: the proxy prediction phase infers the semantic roles of faces and edges, and the proxy-to-asset instantiation phase assembles the proxies into a structured building. Red lines uniformly denote activated edge slots, which assume distinct structural roles in practice.}
    \label{pipeline}
\end{figure*}

First, in the proxy prediction phase, we design a face-edge bigraph encoder to process pre-subdivided, semantic-free building shells comprising only basic geometric primitives. By explicitly propagating and interacting the features of face regions (e.g., walls, roofs) and polylines (e.g., railings, pillars) within the latent space, the network accurately infers the functional roles of each topological element in the generation process. Subsequently, in the proxy-to-asset instantiation phase, we leverage an LLM to parse global styles and procedural attributes from text. These attributes, combined with the geometric features of the primitives, are then utilized to perform asset retrieval for individual proxy slots. Guided by a hard-constrained spatial placement logic, the abstract proxies are precisely instantiated into structured building models. To support the training and evaluation of this pipeline, we construct a building dataset containing textual descriptions, building shells, MAPP semantic labels, and complete building instances built upon prior works \cite{selvaraju2021buildingnet, huang2025buildingblock}. Furthermore, to verify the geometric prior learning and generalization of ProxyBuild, we extend the test data sources to include building meshes generated by BuildAnyPoint \cite{hua2026buildanypoint} and real-world urban LoD2 segments from 3DBAG \cite{Peters22}. Experiments demonstrate that ProxyBuild can perform highly accurate structured parsing and automated asset assembly on these semantic-free meshes, achieving superior performance in terms of generative diversity, structural rationality, and editability. The main contributions of this paper can be summarized as follows:

\begin{itemize}
\item We propose ProxyBuild, a novel hybrid approach that integrates graph neural network-based topological proxy inference, text-conditioned semantic understanding, and procedural content generation, enabling structured building generation jointly driven by text and base geometry.
\item We introduce MAPP as a new building intermediate representation and design a face-edge bigraph encoder. By explicitly modeling the feature interactions between faces and edges on the mesh subdivision graph, this network overcomes the representational limitations of bounding boxes and achieves accurate primitive role prediction.
\item ProxyBuild reduces common geometric artifacts in building generation. Furthermore, its editability provides users with flexible local interactive control, facilitating the efficient generation of diverse, detail-rich buildings.
\end{itemize}

\section{Related Work}

\subsection{Procedural Building Modeling}

PCG aims to automatically synthesize digital content using predefined rules, grammars, or parametric systems. It is widely applied in simulating plant growth \cite{raistrick2023infinite, ghrer2026learning} and generating buildings \cite{parish2001procedural, nishida2018procedural}, offering the key advantages of high efficiency and editability. Notably, the Computer Generated Architecture (CGA) shape grammar \cite{muller2006procedural} and its following work \cite{muller2007image} utilize hierarchical rules to characterize building structures and facade features, thereby demonstrating strong controllability in architectural modeling and urban generation. To alleviate the labor costs associated with manual rule authoring and parameter tuning, recent works \cite{zhou2025scenex, liu2026imagine} have explored integrating LLMs into the procedural modeling pipeline to achieve natural language-guided, urban-scale scene generation. However, these methods struggle to adaptively handle diverse building geometries, and the generated PCG rules are often invalid, leading to topological misalignments and geometric interpenetrations. In contrast, ProxyBuild leverages data-driven graph representation learning, enabling procedural generation to naturally anchor to geometric primitives while refocusing the role of LLMs on high-level semantics and attribute inference, which significantly reduces the reliance on hard-coded rules.

\subsection{Learning-based 3D Generation}

Deep generative models have significantly advanced 3D content generation. Data-driven text and image-to-3D generation methods have evolved from SDS-based optimization \cite{poole2022dreamfusion, tang2023dreamgaussian} to multi-view consistent generation \cite{shi2023mvdream, liu2023syncdreamer}, and further to native 3D latent diffusion models \cite{xiang2026native, lai2025hunyuan3d}, shifting from time-consuming per-instance optimization to second-scale feed-forward generation. Moreover, the generation scope has progressively expanded from single objects to local scenes \cite{yao2025cast, tang2026towards, tang2026zeroscene, chen2026sam}, and even to architectural and city-scale environments \cite{che2026mansion, deng2025citygen, tang2026text2villa, lu2026yo}. These methods excel in open-vocabulary capabilities, visual quality, and general asset creation, effectively addressing the global consistency issues prevalent in procedural approaches. However, they typically aim to produce complete 3D objects or renderable representations. Although recent efforts have explored part-level generation \cite{yang2025omnipart, yan2026x}, they still struggle to provide traceable and editable structural decompositions of objects. ProxyBuild combines neural inference with the assembly of discrete architectural components, yielding building models with explicit hierarchical structures that support downstream editing.

\begin{figure*}
    \centering
    \includegraphics[width=\textwidth]{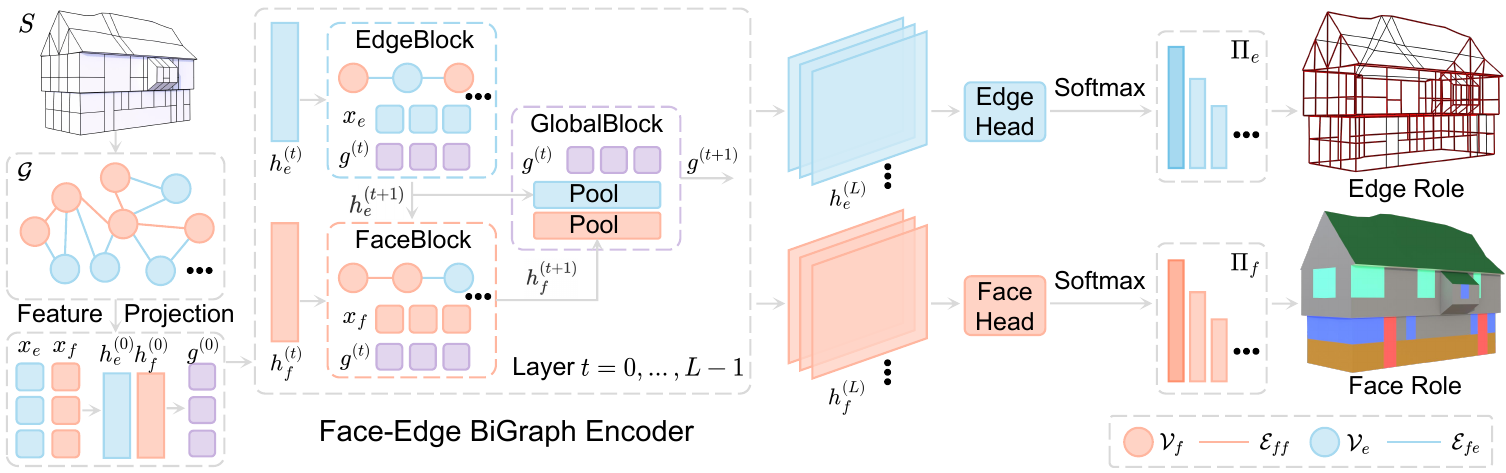}
    \caption{Workflow of the proxy prediction phase and the architecture of the face-edge bigraph encoder. We formulate the semantically-agnostic building shell as a graph structure retaining subdivision information, and conduct feature interaction through specially designed edge, face, and global blocks, ultimately outputting high-precision proxy role probability distributions for both faces and edges.}
    \label{phase1}
\end{figure*}

\subsection{Structured and Hybrid Building Generation}

To balance the structural controllability of procedural modeling and the generative diversity of neural models, recent studies have begun exploring structured and hybrid building generation. The LoD series \cite{tang2025texture2lod3, hanke2025cm2lod3} defines hierarchical representations of building geometry and semantic details. Proc-GS \cite{li2025proc} incorporates procedural rules into the training process of 3DGS, enabling scalable city-level building assembly by manipulating repetitive basic assets such as doors and windows. Pro-DG \cite{plocharski2025pro} recovers hierarchical facade layouts from a single facade image to guide diffusion-based generation, facilitating structured editing such as floor duplication and window rearrangement. BuildingBlock \cite{huang2025buildingblock} trains a diffusion model to generate bounding box layouts for buildings and then employs an LLM to enrich semantic rules and drive procedural modeling. Yet, bounding box representations have inherent limitations in modeling curved walls and complex polyline configurations (e.g., pillars and beams). On the other hand, ShellMaker \cite{xu2026shellmaker} explores language-guided building facade completion on predefined scaffolds via parametric roof generation and component synthesis. Despite these efforts, this method lacks semantic reasoning capabilities for arbitrary geometric shells and is susceptible to distortions from neural part generation. In this paper, we design an intermediate representation based on topological primitives (faces and edges), thereby achieving more natural and fine-grained structural control.
\section{Methodology}

This section first provides a task overview in Section \ref{sec3.1}, followed by detailed descriptions of the proxy prediction phase and the proxy-to-asset instantiation phase in Sections \ref{sec3.2} and \ref{sec3.3}, respectively. The overall pipeline of ProxyBuild is illustrated in Fig. \ref{pipeline}.

\subsection{Overview}\label{sec3.1}

Given a natural language description $T$ and a building shell $S$, our goal is to generate a building model $B$ with rich structural details. Unlike standard LoD2 models that contain only coarse outer contours, the input $S$ is assumed to be a pre-partitioned polygonal mesh. Namely, the potential topological partitions are already reflected in the mesh subdivision, where each face or edge corresponds to an atomic proxy slot, albeit devoid of semantic labels. To bridge the gap between the semantic-free mesh and concrete building assets, we propose the MAPP. Rather than serving as the final geometric entity, MAPP acts as an intermediate control representation bound to the input shell. We formulate it as a graph structure comprising complete probability distributions of proxy roles:
\begin{equation}
    P = \left( \mathcal{G}, \Pi_f, \Pi_e \right), \Pi_f = \{\boldsymbol{\pi}_f\}_{f \in \mathcal{V}_f}, \Pi_e = \{\boldsymbol{\pi}_e\}_{e \in \mathcal{V}_e}
    \label{eq1}
\end{equation}
where $\mathcal{G}$ denotes the face-edge incidence graph; $\boldsymbol{\pi}_f$ and $\boldsymbol{\pi}_e$ represent the proxy role probability distributions for face and edge nodes, respectively, both containing a Null role to filter out background primitives that require no instantiation. Face roles describe planar functionalities such as walls, roofs, and windows, while edge roles characterize linear structures like pillars, beams, and railings. 

To decouple structural plausibility from stylistic diversity, we design a two-stage generative paradigm. First, the Proxy Prediction Phase infers the proxy $P$ from the shell $S$ based on geometric features and topological contexts. Then, guided by the text condition $T$, the Proxy-to-Asset Instantiation Phase converts $P$ into attributed proxies with specific asset mappings and physical parameters, retrieves compatible assets from the asset library $\mathcal{L}_{asset}$ via an adapter, performs spatial pose alignment, and finally assembles them into a structured building model $B$.

\subsection{Proxy Prediction Phase}\label{sec3.2}

As shown in Fig. \ref{phase1}, in this phase, a face-edge bigraph model is constructed to infer the proxy roles of the mesh subdivision primitives.

\paragraph{Subdivision-Preserving Graph Construction} Given a shell $S$, we build a heterogeneous graph $\mathcal{G} = (\mathcal{V}_f \cup \mathcal{V}_e , \mathcal{E}_{fe} \cup \mathcal{E}_{ff})$, where $\mathcal{V}_f$ and $\mathcal{V}_e$ are the sets of face and edge nodes, respectively. A face-edge incidence edge $\mathcal{E}_{fe}$ exists if an edge $e$ belongs to the boundary of a face $f$, while a face-face adjacency edge $\mathcal{E}_{ff}$ connects two faces sharing a common edge $e$. The input geometric features $x_f$ of each face node include center coordinates, normal vector, area, relative building height, and local planarity. The edge node features $x_e$ comprise length, direction, dihedral angle, and boundary status. All spatial features are normalized within the local building coordinate system. Additionally, we record the actual metric geometric parameters $q_i$ of each primitive to preserve absolute physical scale for subsequent retrieval and placement.

\paragraph{Face-Edge BiGraph Encoder} To explicitly capture the interdependencies between faces and edges, we design a dual-stream graph neural network. Since the raw geometric features of faces and edges differ in dimensionality, we first map them to a unified hidden dimension through independent projection networks, $\phi_f$ and $\phi_e$:
\begin{equation}
    h_f^{(0)} = \phi_f(x_f), \quad h_e^{(0)} = \phi_e(x_e)
    \label{eq2}
\end{equation}
The global state $g$ is constructed by aggregating the initialized primitive features and concatenating the bounding box dimensions of the building, ensuring that the network captures the overall shape and relative proportions from the very first layer:
\begin{equation}
\resizebox{0.90\linewidth}{!}{$
    g^{(0)} = \phi_g\left( \text{MeanPool}(\{h_f^{(0)}\}) \parallel \text{MeanPool}(\{h_e^{(0)}\}) \parallel \text{Size}_{\text{bbox}} \right)
$}
\label{eq3}
\end{equation}
At the $t$-th round of message passing, the edge hidden state $h_e$ and the face hidden state $h_f$ are jointly updated via the following mechanisms:
{\footnotesize
\begin{equation}
\begin{aligned}
  h_e^{(t+1)} &= \text{EdgeBlock}\Big(h_e^{(t)}, \text{Agg}_{f \in \mathcal{N}_{fe}(e)}\big(h_f^{(t)}\big), x_e, g^{(t)}\Big) \\
  h_f^{(t+1)} &= \text{FaceBlock}\Big(h_f^{(t)}, \text{Agg}_{e \in \mathcal{N}_{fe}(f)}\big(h_e^{(t+1)}\big), \\
              &\qquad\quad \text{Agg}_{f' \in \mathcal{N}_{ff}(f)}\big(h_{f'}^{(t)}\big), x_f, g^{(t)}\Big) \\
  g^{(t+1)}   &= \text{GlobalBlock}\Big(g^{(t)}, \text{Pool}\big(\{h_f^{(t+1)}\}\big), \text{Pool}\big(\{h_e^{(t+1)}\}\big)\Big)
\end{aligned}
\label{eq4}
\end{equation}
}
where $\mathcal{N}(\cdot)$ denotes the set of neighboring nodes under the corresponding relation, and $\text{Agg}(\cdot)$ is a permutation-invariant aggregation operator. The EdgeBlock enables each edge to aggregate the semantic context from its incident faces; the FaceBlock jointly models the features of contour edges and adjacent faces; and the GlobalBlock updates the global vector $g$ to maintain the macroscopic context. After $L$ layers of interaction, the network outputs class probability distributions $\boldsymbol{\pi}_f$ and $\boldsymbol{\pi}_e$ for faces and edges, respectively, through parallel classification heads.

\paragraph{Training Objective} The joint loss function of the network is defined as:
\begin{equation}
    \mathcal{L} = \mathcal{L}_{\text{face}} + \mathcal{L}_{\text{edge}} + \lambda_s \mathcal{L}_{\text{struct}}
    \label{eq5}
\end{equation}
Here, $\mathcal{L}_{\text{face}}$ and $\mathcal{L}_{\text{edge}}$ employ a class-balanced focal loss to mitigate the class imbalance between dominant elements (e.g., walls) and sparse components (e.g., doors). $\mathcal{L}_{\text{struct}}$ is a differentiable structural consistency regularizer applied to the soft probability outputs of the network, penalizing predictions that violate architectural common sense, thereby enhancing the physical plausibility of the generated structures without requiring additional manual annotations. Additional details are provided in Appendix \ref{app:loss}.

\subsection{Proxy-to-Asset Instantiation Phase}\label{sec3.3}

\begin{figure}
    \centering
    \includegraphics[width=\linewidth]{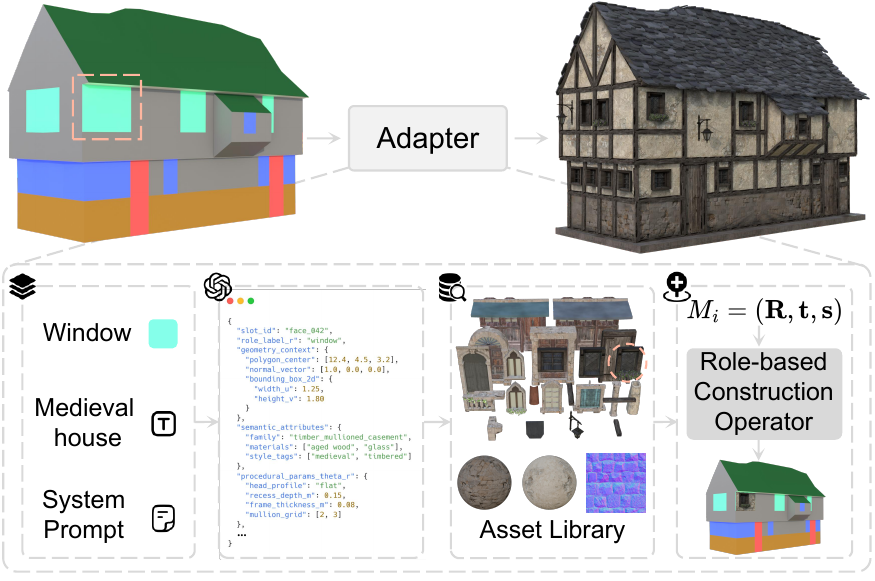}
    \caption{Details of the adapter during the proxy-to-asset instantiation phase. Taking a face-activated slot assigned the "window" role as an example, the adapter leverages the prior knowledge of the LLM to parse global styles and parameters from the text. Subsequently, it retrieves assets by calculating a joint matching score and completes the component placement via hard-constraint geometric solving, finally generating a structured building.}
    \label{phase2}
\end{figure}

After obtaining the structural proxy $P$, the adapter transforms abstract primitive slots into instantiated components equipped with specific asset mappings and attribute parameters. Fig. \ref{phase2} illustrates the workflow of this phase.

\paragraph{LLM-Assisted Attribute Parsing} By filtering out the uninstantiated background roles from $P$, the remaining primitives constitute the set of active slots $\mathcal{I}_{act}$. For each role category, we formulate a prompt (see the complete system prompt in Appendix \ref{app:prompt}) using the global text condition $T$ and the role label $r$, leveraging an LLM to infer the shared procedural geometric parameters $\theta_r$ (e.g., the uniform recess depth of windows or the default cross-sectional radius of pillars):
\begin{equation}
    \theta_r = \text{LLM}\left( \text{Prompt}(T, r) \right)
    \label{eq6}
\end{equation}
Subsequently, for any active slot $i \in \mathcal{I}_{act}$, its predicted role is $r_i = \arg\max \boldsymbol{\pi}_i$. We directly assign it the shared attributes of the corresponding role, which are further converted into physical values $\theta_i$ based on the measured metric features $q_i$. The slot-specific physical dimensions and spatial transformations, on the other hand, are independently computed by the following spatial placement logic.

\paragraph{Constraint-Aware Asset Retrieval and Placement} Based on an active slot $i$ and its predicted parameters $\theta_i$, the adapter computes a multi-dimensional joint matching score for each 3D asset $a$ from the asset library $\mathcal{L}_{asset}$:
\begin{equation}
\resizebox{0.90\linewidth}{!}{$
    \text{Score}(a, i) = \frac{\omega_1 \Phi_{\text{text}}(a, T_i) + \omega_2 \Phi_{\text{geom}}(a, q_i) + \omega_3 \Phi_{\text{param}}(a, \theta_i)}{\omega_1 + \omega_2 + \omega_3}
$}
\label{eq7}
\end{equation}
where $\Phi_{\text{text}}$ calculates the similarity between the asset metadata and the textual style; $\Phi_{\text{geom}}$ evaluates the scale alignment between the asset bounding box and the target slot; and $\Phi_{\text{param}}$ assesses whether the candidate asset supports the predicted procedural parameters. Refer to Appendix \ref{app:score} for the detailed equations.

To mitigate common artifacts in building generation, such as floating components or mesh interpenetrations, the adapter integrates a spatial placement logic based on hard constraints. Concretely, taking the retrieved Top-K highest-scoring assets as input, it first applies geometric adjustments to the candidate assets $a_i$ using the role parameters $\theta_i$ (e.g., updating the collision occupancy boundary of a window according to recess depth). Then, it sequentially solves for a placement by incorporating the topological boundary information of slot $i$ until a feasible solution $a_i^*$ is found, and computes the final physical transformation tuple $M_i = (\mathbf{R}, \mathbf{t}, \mathbf{s})$. During this process, the adapter enforces strict alignment rules: the scaling $\mathbf{s}$ of surface-bound assets must guarantee that their 2D projected contours are strictly enclosed within the target polygons; the translation $\mathbf{t}$ of ground-contact roles is forcefully constrained to align their standard support anchors with the foundation plane. To prevent inter-component geometric penetrations, neighborhood collision detection is performed based on the deformed true bounding boxes. If all Top-K candidates prove infeasible, the system falls back to a default procedural asset. Following the aforementioned parameter prediction and constraint solving, the adapter outputs a packaged set of attributed proxies $\{ (a_i^*, \theta_i, M_i) \}_{i \in \mathcal{I}_{act}}$. Finally, utilizing role-based construction operators (see Appendix \ref{app:operators} for details), the instantiated components are fused with the original shell $S$ to generate the structured 3D building mesh $B$.

\paragraph{Traceable Incremental Editing} Enabled by its stage-decoupled design and explicit graph structure, ProxyBuild natively supports traceable, local incremental editing. When modifying the styles or attributes of selected slots, the system preserves the initial proxy prediction and only re-executes the parameter generation, retrieval, and placement logic for the target slot and its constraint-coupled neighborhood along the graph. If user edits involve changes in roles, geometry, or topology, the system updates the relational graph accordingly and re-runs the complete proxy prediction pipeline. This characteristic affords ProxyBuild a high degree of interactive editing freedom with minimal computational overhead.
\section{Experiments}

\subsection{Experimental Setup}

\paragraph{Datasets} We construct a dataset comprising text descriptions, building shells, MAPP role labels, and complete building instances to train and validate the graph encoder. Specifically, we select instances with full building structures from BuildingNet \cite{selvaraju2021buildingnet}, render their 2D images from 8 different viewpoints, and then employ a multimodal LLM \cite{singh2025openai} to generate independent descriptions for each view, which are further distilled into a globally consistent textual summary. For each building, we extract its semantic-free low-poly shell and back-project the bounding box labels from BuildingBlock \cite{huang2025buildingblock} onto the shell, followed by manual correction to obtain the MAPP annotations. The final dataset consists of 1K instances, encompassing 12 roles, with a total of 126K face annotation slots and 168K edge slots. Additionally, to evaluate out-of-distribution robustness, we expand the test set to clipped segments of real-world 3DBAG \cite{Peters22} city meshes (LoD2) and building geometry produced by the point cloud-to-mesh method BuildAnyPoint \cite{hua2026buildanypoint}.

\begin{table}
    \centering
    \caption{Quantitative comparison of different generation methods.}
    \label{quantitative}
    \resizebox{\columnwidth}{!}{%
    \begin{tabular}{l ccccc}
    \toprule
    \textbf{Method} & \textbf{FID $\downarrow$} & \textbf{KID $\downarrow$} & \textbf{CLIP-S $\uparrow$} & \textbf{UNI3D $\uparrow$} & \textbf{NCR (\%) $\uparrow$} \\
    \midrule
    Meshy & 45.38 & 8.62 & 0.317 & 0.351 & - \\
    Trellis & 52.15 & 10.17 & 0.299 & 0.324 & - \\
    BuildingBlock-Box & 38.52 & 6.51 & 0.341 & 0.385 & 64.3 \\
    ShellMaker & 46.73 & 9.48 & 0.282 & 0.319 & \textbf{96.1} \\
    \textbf{Ours} & \textbf{32.97} & \textbf{5.93} & \textbf{0.349} & \textbf{0.438} & 95.7 \\
    \bottomrule
    \end{tabular}%
    }
\end{table}

\paragraph{Implementation Details} In the proxy prediction phase, the projection networks $\phi_f$ and $\phi_e$ are multi-layer perceptrons (MLPs) that map input features to a hidden dimension of 96. The Face-Edge BiGraph Encoder consists of $L=4$ message-passing interaction layers, and also employs MLPs as parallel classification heads. The model is trained on a single NVIDIA L20 GPU for 31,250 iterations with a batch size of 16, using the AdamW optimizer with an initial learning rate of $1 \times 10^{-3}$. The weight hyperparameters in Eq. \ref{eq5} and Eq. \ref{eq7} are set to $\lambda_s=0.5$, and $\omega_1=1.5$, $\omega_2=1.2$, $\omega_3=0.8$, respectively. In the proxy-to-asset instantiation phase,  the LLM responsible for attribute parsing is GPT-5 \cite{singh2025openai}. The asset library $\mathcal{L}_{asset}$ comprises high-quality models collected from platforms like Sketchfab, alongside multi-stylized components procedurally generated via Blender Geometry Nodes \cite{blender}, all accompanied by metadata including textual descriptions and bounding box dimensions.

\paragraph{Baselines} We compare against two categories of baselines. The first includes general 3D generation methods, Meshy \cite{hu2024meshy} and Trellis \cite{xiang2025structured}. Both approaches directly generate 3D meshes from text and are used to evaluate the overall visual quality. The second comprises structured pipelines, BuildingBlock-Box and ShellMaker \cite{xu2026shellmaker}, used to evaluate the plausibility of structured generation. The former employs BuildingBlock \cite{huang2025buildingblock} to generate bounding box layouts, selects assets via nearest-neighbor retrieval from the same asset library $\mathcal{L}_{asset}$ as ours, and independently scales them to fit the target bounding boxes; the latter combines procedural roof construction, image-to-3D part generation, and joint texture retrieval to perform facade completion on scaffolds.

\begin{table}
    \centering
    \caption{Human preference. All values are the mean scores of 200 independent ratings on a 1-10 scale.}
    \label{userstudy}
    \resizebox{\columnwidth}{!}{%
    \begin{tabular}{l cccc}
    \toprule
    \textbf{Method} & \textbf{Visual Appeal $\uparrow$} & \textbf{Structural Plausibility $\uparrow$} & \textbf{Text Consistency $\uparrow$} \\
    \midrule
    Meshy & 7.1 & 6.7 & 7.9 \\
    Trellis & 6.4 & 6.2 & 7.4 \\
    BuildingBlock-Box & 7.5 & 7.7 & 8.4 \\
    ShellMaker & 7.3 & 7.5 & 7.2 \\
    \textbf{Ours} & \textbf{8.2} & \textbf{8.1} & \textbf{8.5} \\
    \bottomrule
    \end{tabular}%
    }
\end{table}

\paragraph{Evaluation Metrics} We adopt FID \cite{heusel2017gans} and KID \cite{kid} to measure the visual discrepancy between the generated results and the real data distribution, and CLIP Score (CLIP-S) \cite{radford2021learning} to evaluate the semantic consistency between 2D renderings of generated models and text prompts. We further utilize UNI3D \cite{zhou2024uni3d} to directly quantify view-invariant 3D-text alignment in the 3D feature space. To measure structural rationality, we introduce the Non-Collision Rate (NCR) to quantify illegal volumetric intersections (e.g., collisions between windows and walls) among components. Moreover, we report the accuracy of the roles assigned to each topological primitive by the graph encoder, denoted as Proxy Accuracy (ACC).

\begin{figure*}
    \centering
    \includegraphics[width=\linewidth]{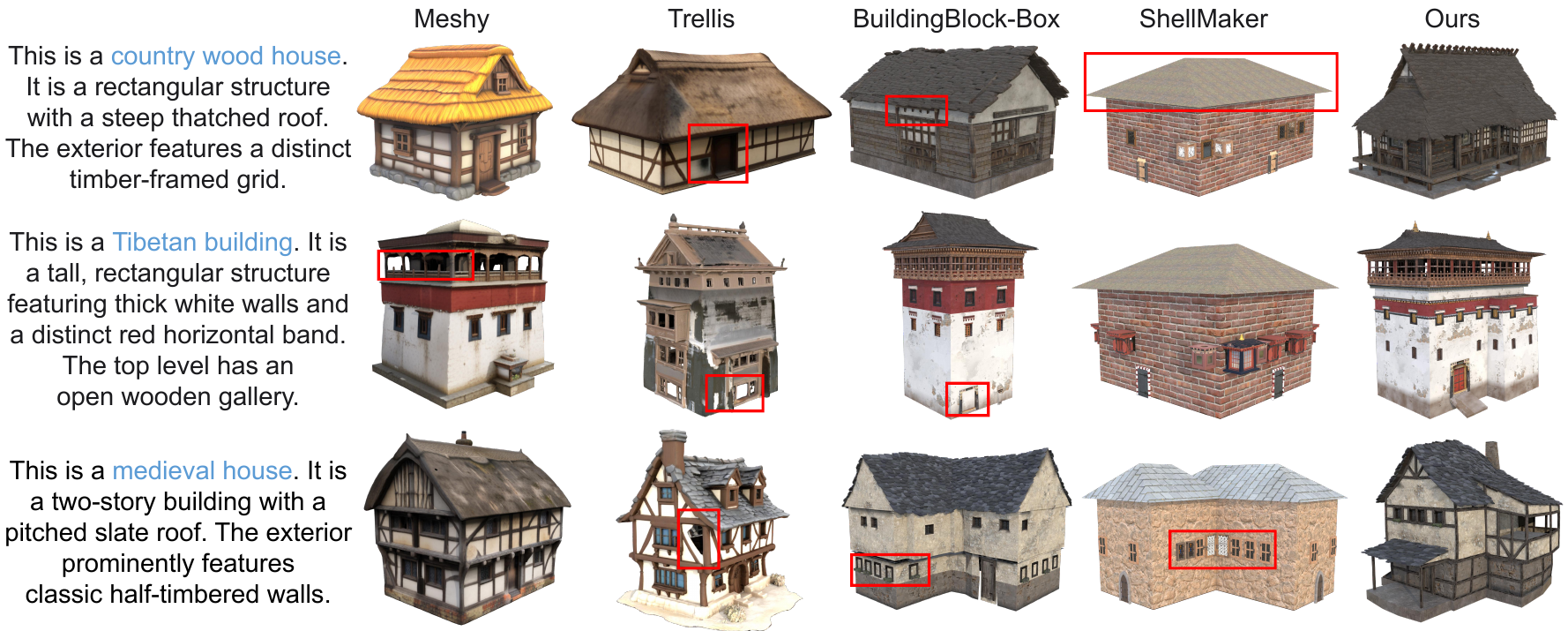}
    \caption{Qualitative comparison of building generation results with state-of-the-art methods. Artifacts are highlighted with red boxes.}
    \label{qualitative}
\end{figure*}

\begin{figure}
    \centering
    \includegraphics[width=\linewidth]{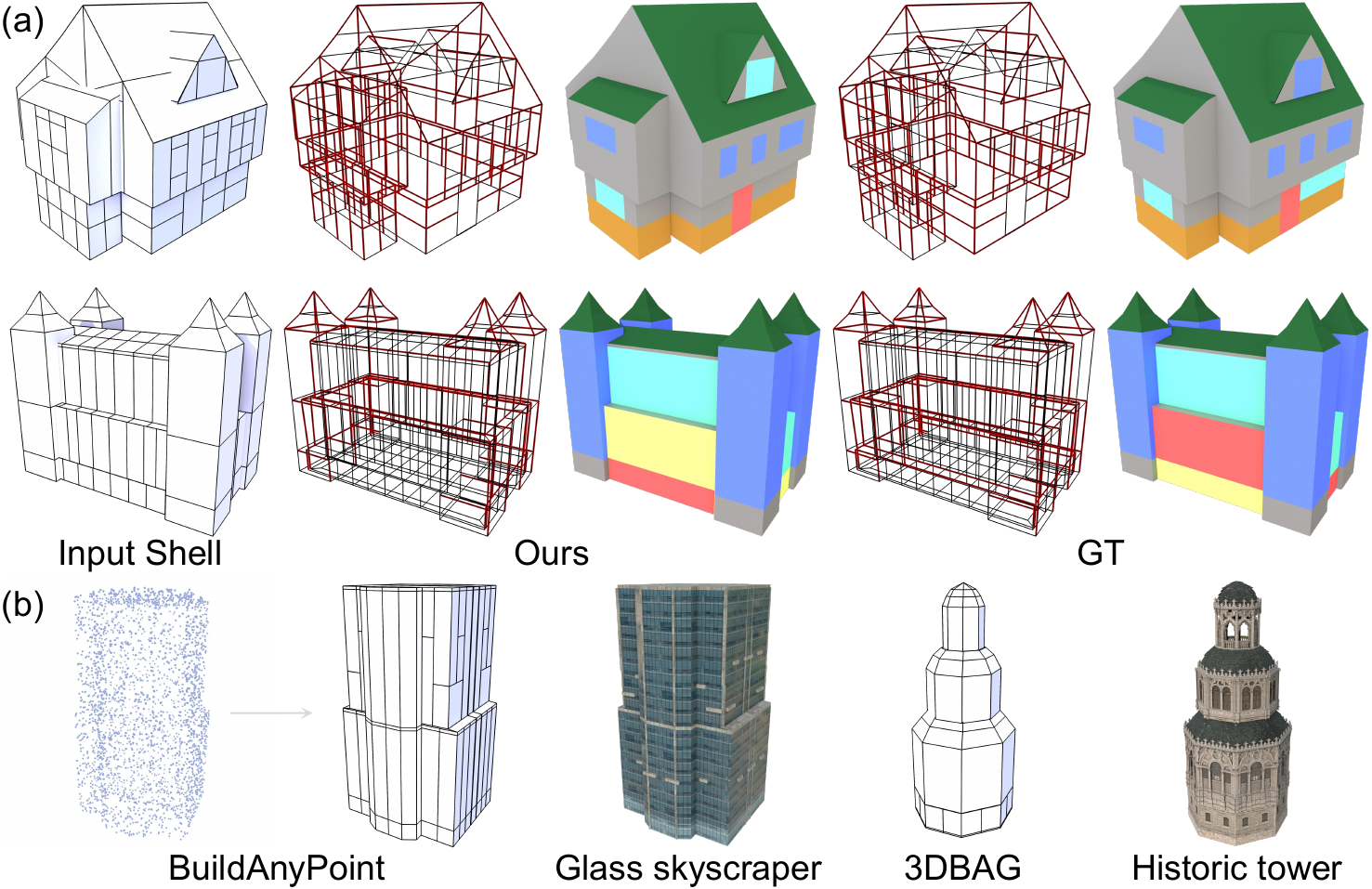}
    \caption{Proxy role prediction and generalization capabilities. (a) Visual comparison between the predicted face and edge roles and ground-truth labels, verifying high predictive accuracy. (b) Generalization to diverse and unseen shells, demonstrating that ProxyBuild can effectively parse point cloud-reconstructed meshes from BuildAnyPoint, as well as real-world urban shells with varying topological densities such as those from 3DBAG.}
    \label{qualitative2}
\end{figure}

\subsection{Experiment Results}

\subsubsection{Quantitative Comparison} Table \ref{quantitative} presents the quantitative results of different methods. Although Meshy and Trellis achieve acceptable visual quality metrics, their outputs are monolithic, unparsable meshes, which prevents meaningful evaluation on structural metrics such as NCR. Compared to BuildingBlock-Box, our method demonstrates significant advantages across all metrics. This is primarily because relying solely on bounding boxes fails to tightly fit complex architectural contours, and forcibly scaling assets often leads to stretching artifacts and interpenetration. In addition, although ShellMaker achieves a high NCR owing to rigid scaffold boolean cuts, its feedforward neural part generation easily induces noise, leading to degraded visual fidelity and low text alignment. In contrast, ProxyBuild maintains high structural plausibility while also leading in visual quality, benefiting from its mesh-anchored proxies and hard-constraint adapters.

Additionally, Table \ref{userstudy} reports the preference scores obtained from a blind evaluation involving 20 human volunteers. Our method achieves overwhelming superiority in both visual appeal and structural plausibility, while also outperforming all baselines in text consistency. The implementation of the user study can be found in Appendix \ref{app:human}.

\paragraph{Qualitative Comparison} Fig. \ref{qualitative} illustrates the qualitative comparisons of generation results across different methods. Meshy often produces buildings with over-smoothed surfaces and a noticeable cartoonish style. Trellis, on the other hand, struggles with local components such as doors and windows, exhibiting artifacts including blurred contours and irregular holes. Although BuildingBlock-Box can generate discrete components, its bounding box representation fails to effectively process curved surfaces and polylines. Consequently, generated windows frequently suffer from spatial misalignments, either sinking into the walls or floating outside, while pillars fail to accurately support the eaves. Despite preserving macro-level volume through strict constraints, ShellMaker is limited by a simple roof generation strategy and isolated component generation, failing to produce fine structural elements representative of styles and resulting in homogenized facades that lack detail. In contrast, buildings generated by ProxyBuild demonstrate superior geometric clarity and well-organized facades, featuring tight and seamless connections between components. Furthermore, Fig. \ref{qualitative2} (a) provides a visual comparison between the face-wise and edge-wise role prediction results and the ground truth. To verify the generalization ability of ProxyBuild across diverse mesh sources, topological densities, and surface partitions, we apply it to 3DBAG LoD2 segments and meshes generated by BuildAnyPoint \cite{hua2026buildanypoint}, with the results shown in Fig. \ref{qualitative2} (b).

\begin{figure*}
    \centering
    \includegraphics[width=\textwidth]{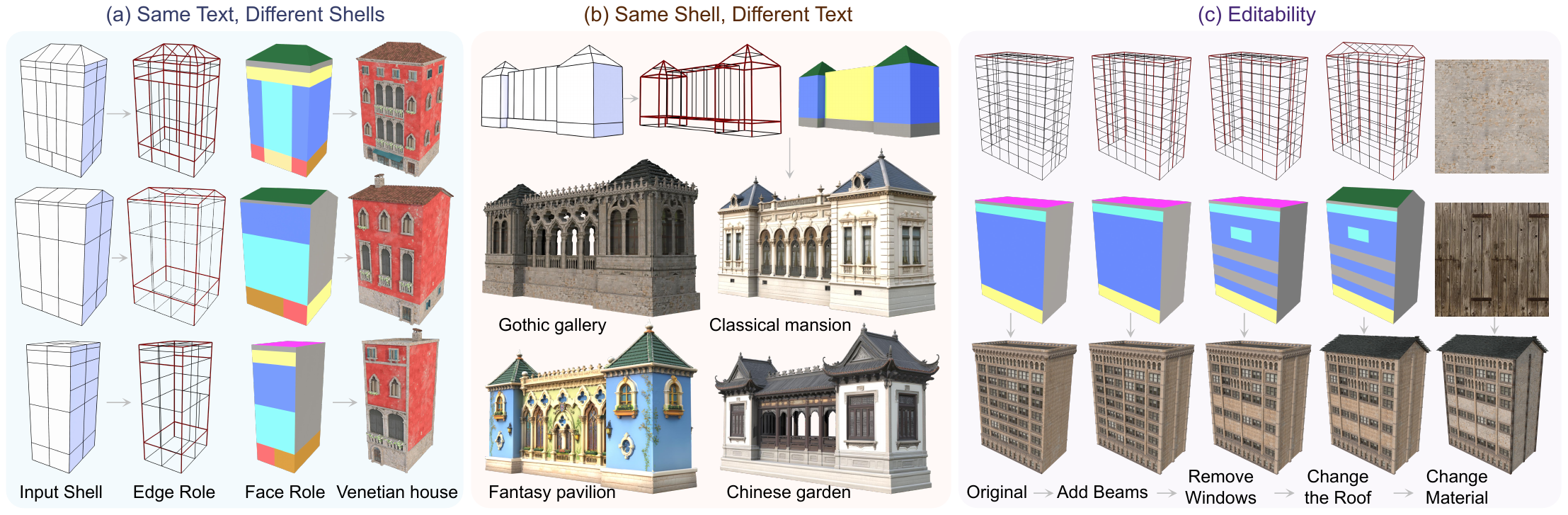}
    \caption{ProxyBuild supports flexible, controllable generation and editing. (a) Identical text guiding diverse shells: adapting to arbitrarily structured meshes to generate buildings with varied morphologies but consistent styles. (b) Identical shell combined with diverse texts: deploying differentiated assets and materials while preserving the macroscopic topology. (c) Progressive editing: allowing users to perform precise local modifications without affecting the global structure.}
    \label{qualitative1}
\end{figure*}

\paragraph{Controllable and Diverse Generation} Empowered by MAPP, ProxyBuild enables flexible and diverse building generation. As illustrated in Fig. \ref{qualitative1}, we demonstrate its controllability through three groups of experiments. 1) Same text, different shells. Given varying building shells and an identical text description, our graph encoder allocates proxy slots in a reasonable manner onto meshes of arbitrary structures, producing buildings with diverse morphologies but a highly consistent style. 2) Same shell, different text. When provided with an identical building shell but different text, the network maintains the consistency of topological roles during the proxy prediction phase, while successfully dispatching differentiated assets and materials during the instantiation phase. The generated buildings exhibit distinctly different styles and facade details while preserving the same macroscopic morphology. 3) Editability. ProxyBuild offers flexible support for user customization. Modifications to text or attributes only require re-adaptation and placement of the corresponding slots. Users can independently add or remove building components, adjust properties such as dimensions, correct the role predictions of specific faces or edges, or directly replace assets and materials. This editing capability facilitates a seamless transition from conceptualization to detailed design, thereby yielding diverse and highly detailed building instances.

\begin{table}
    \centering
    \caption{Quantitative ablation study on the key designs of ProxyBuild.}
    \label{ablation}
    \resizebox{\columnwidth}{!}{%
    \begin{tabular}{l cccc}
    \toprule
    \textbf{Model Variant} & \textbf{CLIP-S $\uparrow$} & \textbf{UNI3D $\uparrow$} & \textbf{NCR (\%) $\uparrow$} & \textbf{ACC (\%) $\uparrow$} \\
    \midrule
    W/o Edge Stream & 0.339 & 0.407 & 92.1 & 76.4 \\
    W/o Hard Constraint & 0.325 & 0.362 & 73.6 & 91.5 \\
    \textbf{Full Model} & \textbf{0.349} & \textbf{0.438} & \textbf{95.7} & \textbf{91.5} \\
    \bottomrule
    \end{tabular}%
    }
\end{table}

\subsection{Ablation Study} We validate the necessity of the key designs in ProxyBuild, with quantitative results summarized in Table \ref{ablation}. The results indicate that removing the edge stream features renders the network incapable of recognizing elongated components attached to boundaries, leading to a significant drop in ACC. Removing the hard constraint module does not affect proxy role prediction but causes NCR to plummet to 73.6\%, resulting in numerous geometric errors of interpenetrating components in the generated outputs. This demonstrates the critical role of incorporating hard constraints for spatial pose correction during the instantiation phase.

\section{Discussion}

\begin{figure}
    \centering
    \includegraphics[width=\linewidth]{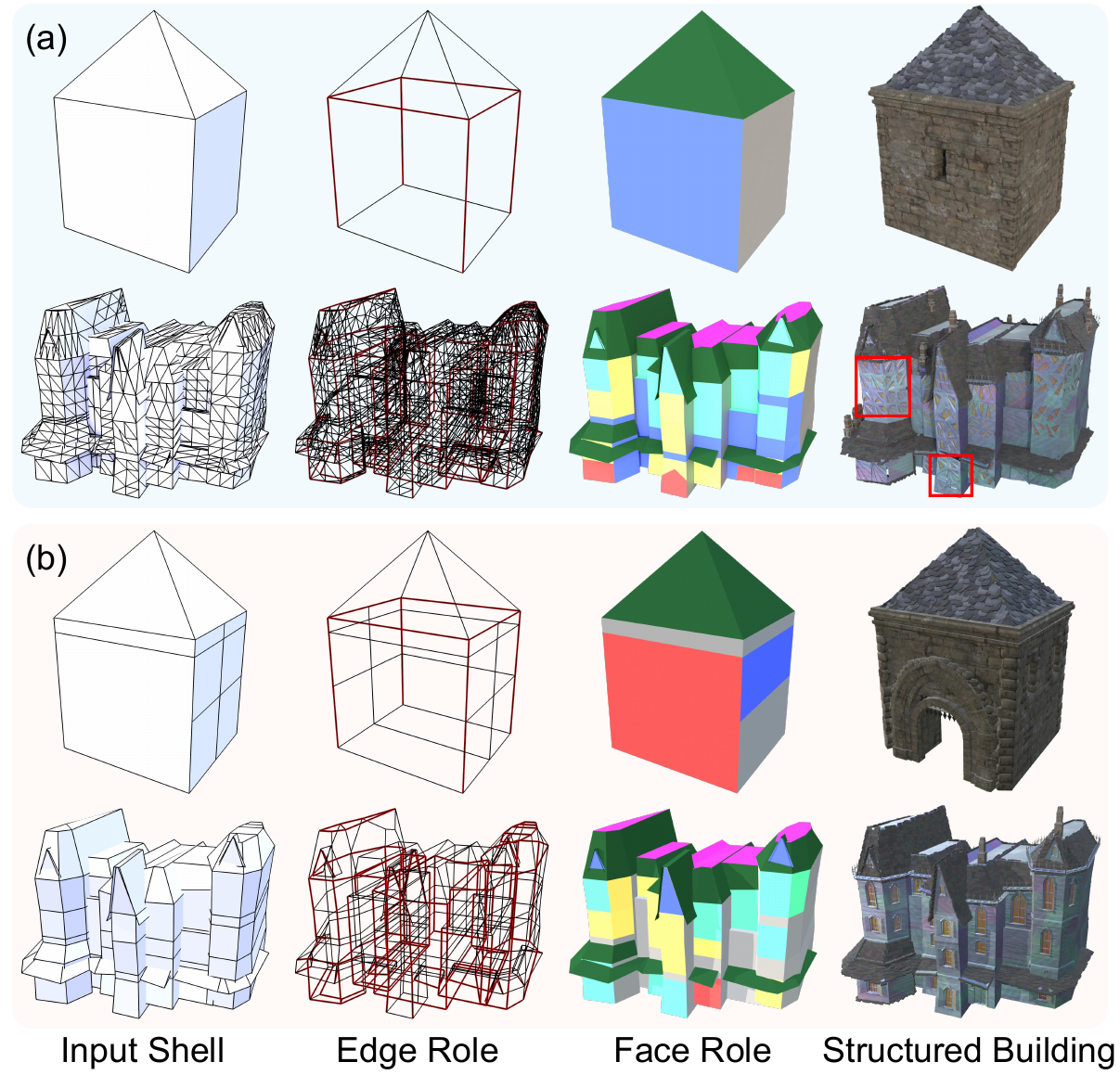}
    \caption{Impact of shell topology and preprocessing. (a) Failure cases: Coarse, large-area face partitioning leads to monotonous structures lacking architectural details, while overly subdivided meshes cause implausible proxy role predictions and severe component distortions. (b) Results with structure-aware remeshing: Regularizing the shell topology into well-proportioned polygonal patches effectively recovers the structural richness and assembly fidelity.}
    \label{failurecase}
\end{figure}

As shown in Fig. \ref{teaser}, ProxyBuild can handle input shells of diverse morphologies. It achieves remarkable progress in structured building generation and opens up several avenues for future exploration. First, the performance of the graph encoder is inherently constrained by the topological validity of the input shell, as demonstrated in Fig. \ref{failurecase} (a). If the polygon partitioning is overly coarse, the generated building facades often lack essential structural hierarchy and semantic diversity. Conversely, an over-tessellated mesh not only substantially increases the computational overhead of graph nodes, but its physical scale also becomes inadequate for effective component instantiation. This issue can be alleviated via a structure-aware remeshing preprocessing pipeline: for large-area facets, adaptive planar subdivision is applied according to the architectural bay scale; for dense and fragmented meshes, the feature-preserving mesh simplification algorithm QSlim \cite{garland1997surface} is first utilized to eliminate coplanar redundant triangles, followed by the remeshing method QuadriFlow \cite{huang2018quadriflow} to extract regular polygonal patches. As presented in Fig. \ref{failurecase} (b), this preprocessing effectively regularizes the facet topology and physical scale, significantly improving the generation quality. Integrating dynamic graph subdivision and pooling mechanisms within the network also offers a potential avenue to alleviate this problem. Second, the current asset retrieval step relies on a finite offline database, which restricts its mapping capability when encountering unseen or rare textual styles. Future work could integrate the on-the-fly generation of native 3D models to enable dynamic, open-vocabulary asset expansion. Finally, leveraging the ability of ProxyBuild to automatically synthesize text-structured building data pairs, these rich priors could be utilized to train a dedicated, end-to-end feed-forward model for high-quality building generation. This would forge a highly promising new path for 3D city-scale content generation.

\section{Conclusion}

This paper presents ProxyBuild, a hybrid generation framework for 3D structured buildings jointly driven by text and base geometry. To address the lack of hierarchical structure in existing outputs and the inadequate boundary constraints of traditional bounding box representations, we innovatively introduce MAPP, firmly anchoring the generation task to the geometric topology of the building shell. Through a carefully curated dataset and a face-edge bigraph encoder, ProxyBuild overcomes the semantic parsing bottleneck of polygonal meshes, achieving precise inference of face and edge roles. Building upon this, by integrating textual attributes parsed by an LLM with constraint-aware spatial placement logic, the abstract topological proxies are seamlessly transformed into structurally rigorous and highly detailed instantiated 3D assets. Experiments demonstrate that ProxyBuild successfully breaks down the barrier between unstructured geometric representations and parametric component entities. It not only effectively mitigates artifacts such as spatial misalignment of components and structural fragmentation, but also empowers users with a high degree of free interactive control. ProxyBuild offers an efficient new paradigm for the automated generation of high-quality, editable 3D buildings, laying a solid foundation for future large-scale digital twins and urban-scale content generation.

{
    \small
    \bibliographystyle{ieeenat_fullname}
    \bibliography{main}
}

\clearpage 
\appendix 

\twocolumn[
  \begin{center}
    {\Large \bf ProxyBuild: Text-Guided Structured 3D Building Generation with Mesh-Anchored Procedural Proxies \par} 
    \vspace{1em} 
    {\Large \bf Supplementary Material \par} 
    \vspace{1.5em} 
  \end{center}
]

\section{Network and Training Details}

\subsection{Definition of Face and Edge Role Spaces}

In the graph $\mathcal{G}$, face nodes and edge nodes possess independent semantic role spaces. The face role space $\mathcal{R}_f$ comprises surface classes (\texttt{Empty Wall}, \texttt{Floor}, \texttt{Flat Roof}, \texttt{Sloped Roof}) and instantiated classes (\texttt{Opening}, \texttt{Window}). The edge role space $\mathcal{R}_e$ consists of linear instantiated classes (\texttt{Pillar}, \texttt{Beam}, \texttt{Railing}, \texttt{Diagonal Brace}, \texttt{Pipe}) and a \texttt{Null} role. Each primitive is assigned a single role; consequently, the output probability distributions of the network, $\boldsymbol{\pi}_f$ and $\boldsymbol{\pi}_e$ both reside within the corresponding probability simplices.

\subsection{Loss Function}\label{app:loss}

In Eq. \ref{eq5}, we adopt a class-balanced focal loss for node types $\tau \in \{f, e\}$ as follows:
\begin{equation}
    \mathcal{L}_\tau = -\frac{1}{|\mathcal{V}_\tau|} \sum_{i \in \mathcal{V}_\tau} \alpha^\tau_{y_i} (1 - \pi_{i,y_i})^\gamma \log(\pi_{i,y_i} + \epsilon)
    \label{eq8}
\end{equation}
where $y_i$ is the ground-truth class label for node $i$, and $\pi_{i,y_i}$ represents the predicted probability for this class. The term $\epsilon = 10^{-6}$ is used to maintain numerical stability, and the focusing parameter $\gamma$ is set to 2.0. $\alpha^\tau_{y_i}$ denotes the effective sample number weight, defined as $\alpha^\tau_c = \frac{1-\beta}{1-\beta^{n^\tau_c}}$, where $n^\tau_c$ is the total number of samples for class $c$ in the training set. The decay factor $\beta$ is empirically set to 0.999 to mitigate the class imbalance among minority components.

The structural consistency regularization, $\mathcal{L}_{\text{struct}}$, injects common sense priors into the network by penalizing role combinations that violate architectural plausibility:
\begin{equation}
\begin{split}
    \mathcal{L}_{\text{struct}} &= \frac{1}{\max(1, |\mathcal{E}_{fe}|)} \sum_{(f,e) \in \mathcal{E}_{fe}} \boldsymbol{\pi}_f^\top C_{fe} \boldsymbol{\pi}_e \\
    &+ \frac{1}{\max(1, |\mathcal{E}_{ff}|)} \sum_{\{f,f'\} \in \mathcal{E}_{ff}} \boldsymbol{\pi}_f^\top C_{ff} \boldsymbol{\pi}_{f'}
\end{split}
\label{eq9}
\end{equation}
where $C_{fe} \in \mathbb{R}_{\ge 0}^{|\mathcal{R}_f| \times |\mathcal{R}_e|}$ and $C_{ff} \in \mathbb{R}_{\ge 0}^{|\mathcal{R}_f| \times |\mathcal{R}_f|}$ are predefined penalty cost matrices. Compatible role combinations incur a cost of 0, whereas incompatible ones are assigned a positive penalty. The core construction logic follows two principles. 1) Topological support constraints: For instance, a \texttt{Flat Roof} connected to an edge representing a \texttt{Pillar} or \texttt{Empty Wall} yields a cost of 0, but with a \texttt{Pipe} incurs a cost of 1.0. 2) Coplanar mutual exclusion constraints: For example, if a \texttt{Window} and an \texttt{Opening} are adjacent (i.e., sharing an edge) within the same polygon, a severe penalty is imposed. Penalty values in both rows and columns corresponding to the \texttt{Null} role are uniformly set to 0.

\tcbset{
  llmprompt/.style={
    colback=cyan!6,
    colframe=cyan!60,
    coltitle=black,
    fonttitle=\bfseries,
    boxrule=0.5pt,
    arc=2mm,
    top=1mm,
    bottom=1mm,
    left=2mm,
    right=2mm
  }
}
\begin{figure*}[!ht]\centering
\vspace{-2mm}
\captionof{table}{System prompt for inferring role-shared parameters from a building shell slot.}
\begin{minipage}{\textwidth}\vspace{0mm}    \centering
\begin{tcolorbox}[llmprompt]
\small

You are an expert architectural computational designer. Your task is to process a semantic-free building shell slot. You will use the User's Global Text Prompt and the provided measured Geometry Context to infer precise Semantic Attributes and Procedural Parameters.

\vspace{2mm}

\textbf{[INPUT CONTEXT]}
\begin{itemize}
    \item Global Text Prompt: ``\{User\_Global\_Text\}'' (e.g., ``A medieval house with classic half-timbered walls'')
    \item Slot ID: ``\{slot\_id\}''
    \item Predicted Role: ``\{role\_label\_r\}''
    \item Geometry Context (Real-world metrics of the target slot):
    \begin{itemize}
        \item \texttt{polygon\_center}: \{center\_xyz\}
        \item \texttt{normal\_vector}: \{normal\_xyz\}
        \item \texttt{bounding\_box\_2d}: \texttt{width\_u}=\{width\}, \texttt{height\_v}=\{height\} (in meters)
    \end{itemize}
\end{itemize}

\vspace{2mm}

\textbf{[ASSET LIBRARY CAPABILITIES \& OUTPUT RULES]} \\
Analyze the input and output a strict JSON object. You must retain the \texttt{"slot\_id"}, \texttt{"role\_label\_r"}, and \texttt{"geometry\_context"}. Generate the following two core structures based on our Asset Library limits:

\begin{enumerate}
    \item \texttt{"semantic\_attributes"}:
    \begin{itemize}
        \item \texttt{"family"}: Infer the specific architectural component family (e.g., \texttt{"timber\_mullioned\_casement"}).
        \item \texttt{"materials"}: List 1-3 physical materials.
        \item \texttt{"style\_tags"}: List 2-3 visual keywords for asset retrieval.
    \end{itemize}
    \item \texttt{"procedural\_params\_theta\_r"}:
    \begin{itemize}
        \item MUST be tailored to the Geometry Context metrics. All values MUST be in meters.
        \item If role is \texttt{"window"}: Supported keys are \texttt{"head\_profile"} (flat/arched/pointed), \texttt{"recess\_depth\_m"} (float, typically 0.05-0.40m), \texttt{"frame\_thickness\_m"} (float), \texttt{"mullion\_grid"} (list of 2 ints: [horizontal\_panes, vertical\_panes] calculated based on width/height ratio).
        \item If role is \texttt{"pillar"}: Supported keys are \texttt{"cross\_section"} (circular/square), \texttt{"radius\_or\_width\_m"} (float).
    \end{itemize}
\end{enumerate}

Output ONLY valid JSON matching the exact schema required. Do not include markdown formatting.

\vspace{2mm}

\textbf{[EXAMPLE EXPECTED OUTPUT]}
\vspace{-2mm}
{\scriptsize
\begin{verbatim}  
{
  "slot_id": "face_042",
  "role_label_r": "window",
  "geometry_context": {
    "polygon_center": [12.4, 4.5, 3.2],
    "normal_vector": [1.0, 0.0, 0.0],
    "bounding_box_2d": {
      "width_u": 1.25,
      "height_v": 1.80
    }
  },
  "semantic_attributes": {
    "family": "timber_mullioned_casement",
    "materials": ["aged wood", "glass"],
    "style_tags": ["medieval", "timbered"]
  },
  "procedural_params_theta_r": {
    "head_profile": "flat",
    "recess_depth_m": 0.15,
    "frame_thickness_m": 0.08,
    "mullion_grid": [2, 3]
  }
}
\end{verbatim}
}

\end{tcolorbox}
\end{minipage}
\label{prompt}
\vspace{-4mm}
\end{figure*}

\section{Parsing, Retrieval and Construction}

\subsection{System Prompts for Attribute Parsing}\label{app:prompt}

In the instantiation phase, we leverage an LLM as the attribute parsing engine. Since the model cannot directly process 3D mesh data, we pre-extract the metric features $q_i$ of the target slot $i$ through a script and concatenate them with the global text description as the input context. Meanwhile, to reduce hallucinations and ensure the output parameters are structured for downstream computation, we hardcode the metadata capability priors of each role asset into the prompt. The complete system prompt template is presented in Table \ref{prompt}.

The parsed \texttt{semantic\_attributes} are subsequently used to compute the text semantic alignment score $\Phi_{\text{text}}$. Under the constraints of the slot metric features $q_i$, the shared parameters \texttt{procedural\_params\_theta\_r} (i.e., $\theta_r$) are converted into physical parameters $\theta_i$, which are directly involved in calculating the parameter support score $\Phi_{\text{param}}$ and the subsequent asset instantiation process.

\begin{table*}[t] 
    \centering
    \caption{Construction operators for different roles.}
    \label{operators}
    \renewcommand{\arraystretch}{1.4} 
    \begin{tabular}{m{0.18\textwidth} m{0.18\textwidth} m{0.62\textwidth}}
    \toprule
    \textbf{Role Category} & \textbf{Operator} & \textbf{Geometric Operation \& Constraints} \\
    \midrule
    
    \texttt{Empty Wall}, \newline
    \texttt{Floor}, \newline
    \texttt{Flat Roof}, \newline
    \texttt{Sloped Roof}
    & \texttt{SurfaceReplace}, \newline \texttt{MaterialOverlay} 
    & Maintain the 2D contour of the polygonal slot, forcing the replacement of the original mesh with detailed surface assets (e.g., tiles, wooden boards), or applying material overlays. The 2D bounding contour of the instantiated component is constrained to be strictly contained within the target polygonal domain: $\text{Footprint}_{2D}(M_i^a) \subseteq \Omega_i$. \\
    
    \texttt{Window}, \newline
    \texttt{Opening} 
    & \texttt{InsetAttach} 
    & Recess into the wall interior based on the predicted \texttt{recess depth} parameter. Perform a Boolean Difference or polygonal clipping to cut a hole in the wall surface, and subsequently embed the component into the opening. \\
    
    \texttt{Pillar} 
    & \texttt{SupportAttach} 
    & Align the support anchor point of the component with the underlying load-bearing structure (\texttt{Floor}), and apply the cross-sectional radius predicted by the LLM. The bottom of load-bearing elements, such as \texttt{Pillars}, is constrained to fit the support plane precisely: $\text{dist}(M_i p_a^{\text{sup}}, \Sigma_i^{\text{sup}}) \leq \epsilon_s$. \\
    
    \texttt{Beam}, \newline
    \texttt{Pipe}, \newline
    \texttt{Diagonal Brace}
    & \texttt{EdgeAttach}, \newline \texttt{Sweep} 
    & Place linear components tangentially along the target boundary line, or generate continuous elements by sweeping a given 2D cross-section asset along the boundary edge. \\
    
    \texttt{Railing} 
    & \texttt{RepeatAlong} 
    & Read the default unit length from the asset metadata, perform procedural array instantiation along the topological boundary line, and automatically handle truncation and splicing at corner endpoints. \\

    \texttt{Null} 
    & \texttt{Skip} 
    & Keep the original input primitives intact without any component instantiation. \\
    
    \bottomrule
    \end{tabular}
\end{table*}

\subsection{Multi-dimensional Joint Matching Score}\label{app:score}

To retrieve the most suitable 3D asset from the asset library $\mathcal{L}_{asset}$ for an activated slot $i$, the adapter computes a joint score across three dimensions: semantic, geometric, and parametric. The specific formulation of each term in Eq. \ref{eq7} is detailed as follows.

The text semantic alignment score $\Phi_{\mathrm{text}}$ uses a pre-trained CLIP text encoder $E_{\mathrm{text}}$ (CLIP-ViT-L/14) to compute the cosine similarity between the role \texttt{style\_tags} $T_i$, extracted by the LLM from the global text condition $T$, and the asset metadata $m_a^{\mathrm{text}}$, and normalizes the result to the interval $[0,1]$:
\begin{equation}
    \Phi_{\mathrm{text}}(a, T) = \frac{1 + \cos(E_{\mathrm{text}}(T_i), E_{\mathrm{text}}(m_a^{\mathrm{text}}))}{2}
    \label{eq10}
\end{equation}
The geometric and scale matching score $\Phi_{\text{geom}}$ evaluates the degree of geometric distortion caused by adapting the asset's original bounding box dimensions $d_a$ to the target slot dimensions $d_i$ (provided by $q_i$). To prevent excessive stretching of the components, we compute the ideal per-axis scaling factor vector $s^* = d_i \oslash d_a$ and penalize extreme anisotropic scaling (i.e., severe aspect ratio distortion):
\begin{equation}
\begin{split}
    d_{\text{geom}}(a, i) &= \max\left(0, \ln\frac{\max(s^\star)}{\min(s^\star)} - \ln\kappa_{\text{max}}\right) \\
    \Phi_{\mathrm{geom}}(a, q_i) &= \exp(-d_{\mathrm{geom}}(a, i) / \tau_g)
\end{split}
\label{eq11}
\end{equation}
where $\kappa_{\text{max}} = 1.5$ is the maximum allowable threshold for the non-uniform scaling ratio, and $\tau_g = 0.5$ is a temperature coefficient. The parametric support score $\Phi_{\mathrm{param}}$ assesses whether the asset possesses the procedural adjustability predicted by the LLM:
\begin{equation}
    \Phi_{\mathrm{param}}(a, \theta_i) = \exp\left( -\| D_{r_i}^{-1} (\theta_{i,a} - \theta_i) \|_2^2 \right)
    \label{eq12}
\end{equation}
where $\theta_{i,a}$ is the closest value the asset can provide relative to the expected parameter $\theta_i$, and $D_{r_i}$ denotes the normalizing diagonal matrix for the corresponding parameters.

\subsection{Role-Conditioned Construction Operators}\label{app:operators}

After retrieving the Top-K assets and completing the hard constraint space solution to obtain the final transformation parameter tuple $M_i$, the adapter invokes specific construction operators to fuse the assets into the building shell. In Table \ref{operators}, $\operatorname{Footprint}_{2D}(\cdot)$ denotes the 2D orthogonal projection of the component onto the plane of the target face, and $\Omega_i$ represents the valid 2D boundary of the target polygonal slot within the input mesh. Additionally, $p_a^{\text{sup}}$ is the bottom support point of the 3D asset, $\Sigma_i^{\text{sup}}$ is the underlying load-bearing surface, $\operatorname{dist}(\cdot, \cdot)$ calculates the minimum Euclidean distance, and $\epsilon_s$ is a minimal distance tolerance parameter, set to $10^{-3}$.

During the execution of these construction operators, the adapter follows a prescribed sequence: surface replacement, opening cutting, support component alignment, and finally the attachment of linear and decorative elements. Following each instantiation, the local spatial occupancy is updated based on the resulting bounding box, which enables ProxyBuild to mitigate geometric artifacts such as mesh interpenetrations while ensuring diversity.

\section{Human Preference}\label{app:human}

\begin{figure*}[t!]
    \centering
    \includegraphics[width=\textwidth]{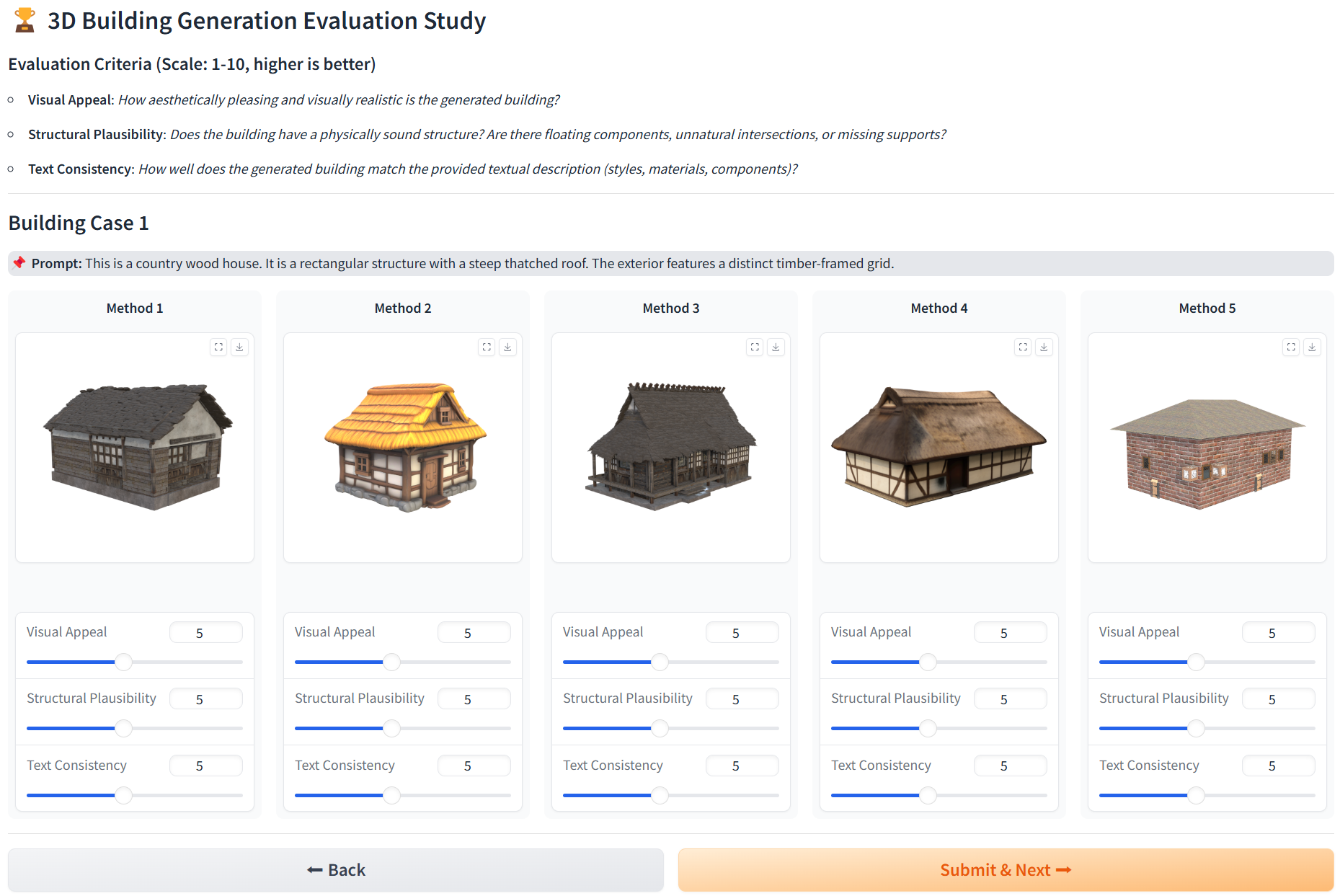}
    \caption{User interface for the human preference study.}
    \label{interface}
\end{figure*}

To comprehensively evaluate the visual and structural quality of the generated buildings from the perspective of human perception, we conduct a user study. We invite 20 volunteers to subjectively evaluate 10 groups of 3D building models via a custom web-based interface shown in Fig. \ref{interface}. For each case, the given text description is displayed at the top of the interface, while the generated results from four baseline methods and ProxyBuild are presented side-by-side below. To eliminate potential bias introduced by the presentation order, the arrangement of the rendered results is randomized for each trial. Participants are asked to carefully observe and then rate each model on a scale of 1 to 10 (where higher scores indicate better performance) across three metrics: visual appeal, structural plausibility, and text consistency. Explicit definitions for each metric are provided within the interface to ensure consistent evaluation criteria among participants. In total, we collect 200 sets of independent and valid rating data. The detailed comparative scores are summarized in Table \ref{userstudy}.

\end{document}